# MODIFIED ALPHA-ROOTING COLOR IMAGE ENHANCEMENT METHOD ON THE TWO-SIDE 2-D QUATERNION DISCRETE FOURIER TRANSFORM AND THE 2-D DISCRETE FOURIER TRANSFORM

Artyom M. Grigoryan [1], Aparna John [1], Sos S. Agaian [2]

[1]University of Texas at San Antonio, San Antonio,TX 78249, USA
[2] City University of New York / CSI

## ABSTRACT

*Color in an image is resolved to 3 or 4 color components and 2-Dimages of these components are stored in separate channels. Most of the color image enhancement algorithms are applied channel-by-channel on each image. But such a system of color image processing is not processing the original color. When a color image is represented as a quaternion image, processing is done in original colors. This paper proposes an implementation of the quaternion approach of enhancement algorithm for enhancing color images and is referred as the modified alpha-rooting by the two-dimensional quaternion discrete Fourier transform (2-D QDFT). Enhancement results of this proposed method are compared with the channel-by-channel image enhancement by the 2-D DFT. Enhancements in color images are quantitatively measured by the color enhancement measure estimation (CEME), which allows for selecting optimum parameters for processing by thegenetic algorithm. Enhancement of color images by the quaternion based method allows for obtaining images which are closer to the genuine representation of the real original color.*

## 1. INTRODUCTION

All color images consist of 3 or 4 channels of monochromatic images. The color of the image is resolved to color components, and pixel values in each channel are the corresponding color image intensities. For example in the RGB color model, the image is stored as three monochromatic images for the red, green, and blue colors. The original color in the image is the color obtained by superposing or merging all three color components. The processing of color images could only be done with fairness if the color of the image is considered as the sum of the color component pixel values in all channels when taken together. But until recently, the processing of the color image was done by splitting up the color image channel-by-channel to individual images and, then, processing each channel with the respective algorithm. To know the totality of enhancement effects of the respective algorithm onto the color image, each enhanced channel is taken together to compose the resulting enhanced image. Ell's work of quaternion approach [11] of color image processing enables us to represent the color of the image as a sum of the color components. Quaternion numbers are four-dimensional hyper-complex numbers. Each of the resolved color components in the pixel can be represented as the component vector in four-dimensional space. That is, a color image is represented as a quaternion image. This representation helps in processing color image by taking the color of the image as the vector sum of color components. The color of the image is considered as a single entity, rather than considering it as three separate resolved color components. In this quaternion approach of color image processing, all three or four channels in the color image are processed simultaneously. This type of enhancement method gives a merging of all of the color components.

Many frequency domain based enhancement algorithms, particularly the Fourier transform-based enhancement algorithms,[1]-[3], [26]-[29] are effective on grayscale images and on color images, when processing channel-by-channel. But there is a need to have efficient color image enhancement algorithms, in which the color is taken as a single entity, or in other words when the processing is done on all channels simultaneously. Many quaternion transform based enhancement methods, which are based on the QDFT, give effective enhancement results. Since quaternions are four-dimensional numbers, the QDFT are transforms in four-dimensional space. Therefore, in orderto implement an enhancement algorithm with the QDFT, there is a need to Figure out fast algorithms. The proposed fast algorithm[2,3,19,30] of QDFT reduces the computational complexities in enhancement algorithms based on the QDFT. The Fourier transform-based enhancement algorithms, such as the alpha-rooting, log alpha-rooting, and modified alpha-rooting methods,which are proven to be effective on grayscale images, can be extended to the quaternion-based color image enhancement. The proposed algorithm is the modified alpha-rooting method followed by different spatial transformation techniques, and the preliminary experimental results show that the proposed enhancement algorithm is effective.

Choice of variables needed for the enhancement algorithm could be made easier by using an optimization algorithm. In order to use such an algorithm, we need to get a cost function that would give a quantitative measure of enhancement in the processed image. Grigory an and Agaian proposed a quantitative measure for enhancements in the enhanced images after processing with the respective enhancement algorithm. The quantitative measure is referred to as the color enhancement measure estimation (CEME). The CEME value is used as the cost function in the proposed optimization algorithm, which is chosen to be the genetic algorithm. This paper describes the method and experimental results of the enhancement by the modified alpha-rooting methods by the 2D-QDFT on color images. Color image processing is done by the quaternion approach, in which all channels in the image are processed simultaneously. The experimental results obtained by the quaternion approach are compared with results obtained by channel-by-channel processing, by the same method of modified alpha-rooting, but with the traditional 2-D DFT. Enhancement in images is measured quantitatively by the CEME value and the optimization of variables is done with the genetic algorithm.

## 2. METHODOLOGY OF 2-DQDFT AND 2-DDFT MODIFIED ALPHA-ROOTING

### 2.1. Quaternion Numbers

Quaternion numbers are four-dimensional hyper-complex numbers[9][10] . A quaternion number $q$ is represented in Cartesian form as

$$q = a + ib + jc + kd. \tag{1}$$

That is, every quaternion number has a scalar part, $S(q) = a$and a vector part, $V(q) = ib + jc + kd$. The imaginary units $i, j, k$ are related as

$$i^2 = j^2 = k^2 = ijk = -1; \tag{2}$$
$$ij = -ji = k \, ; jk = -kj = i; \, ki = -ik = j.$$

The basis $\{1, i, j, k\}$ is the most common and classical basis to express a quaternion. There are many different choices for $q$. Given two pure unit quaternions μ and ψ, that is, $μ^2 = ψ^2 = -1$, which are orthogonal to each other, that is, $\mu \perp \psi$, the set $\{1, \mu, \psi, \mu\psi\}$ is a basis for any quaternion $q$.

The norm of the quaternion defined as

$$\|q\| = a^2 + b^2 + c^2 + d^2. \tag{3}$$

When $\|q\| = 1$, q is said to be unit quaternion. When the scalar part, $S(q) = 0$, then the quaternion is referred as a pure quaternion.The polar form of a quaternion number is given as

$$q = |q|e^{\mu_q \varphi_q} = |q|(cos\,\varphi_q + \mu_q\,sin\,\varphi_q), \tag{4}$$

where$|q|$ is the modulus and $\mu_q$ is called the axis and $\varphi_q$ is the phase or argument of $q$that are expressed as

$$|q| = \sqrt{\|q\|} = \sqrt{(a^2 + b^2 + c^2 + d^2)}, \qquad \mu_q = \frac{\boldsymbol{i}b + \boldsymbol{j}c + \boldsymbol{k}d}{\sqrt{(b^2 + c^2 + d^2)}}, \tag{5}$$

$$\varphi_q = \tan^{-1}\left(\frac{\sqrt{(b^2 + c^2 + d^2)}}{a}\right).$$

An important property in quaternion algebra is the product of two quaternions, whichis not always commutative. That is, for a given two quaternion numbers $p$ and $q$.

$$pq = qp \quad \text{or} pq \neq qp. \tag{6}$$

## 2.2.Color Image In The Quaternion Space

The color image can be represented as a quaternion image [9],11]-[16]. Depending on the color model, the color images have three or four channels. In the case of three channel color models like the RGB, XYZ, or in luminance-chrominance color model $YC_bC_r$, the color images can be represented as a pure quaternion. For example, the color image $f\,(n,m)$ in RGB color model can be represented as a quaternion image

$$f(n,m) = i\,r(n,m) + j\,g(n,m) + k\,b(n,m), \tag{7}$$

where $(n,m)$, $g(n,m)$,and $b(n,m)$ are the red, green and blue components, respectively. In general, for three or four channel color images, the parts$a, b, c,$ and $d$of the quaternion number $q$ take the form[56]

$$a = f_{n,m_{Channel\ 1}}, b = f_{n,m_{Channel\ 2}}, c = f_{n,m_{Channel\ 3}}, d = f_{n,m_{Channel\ 4}}. \tag{8}$$

When a color image is represented as quaternion, we get a correlation between all color components. In the usual color image enhancements, this correlation between the color components is not used, because each channel is processed separately. The even component, $\boldsymbol{a}$ of the quaternion color image can be chosen in many different ways. The two common choices are $a = 0$and the grayscale value, $a = \frac{1}{3}[r + g + b]\boldsymbol{.}$

## 2.3 Two-Side 2-Dquaternion Fourier Transform

The discrete Fourier transform of 2-D hyper-complex numbers is referred to as the 2-D QDFT. As mentioned before, the product of two quaternions are not always commutative. Thus, based on the position of the kernel of the transform with respect to the image, there are many versions of 2D QDFT. In the two-side 2D-QDFT, the image is sandwiched by the two transform kernels and defined as

$$QDFT, F(p,s) = \sum_{n=0}^{N-1} W_j^{np} \left[ \sum_{m=0}^{M-1} f_{n,m} W_k^{ms} \right], \quad p = 0:(N-1), s = 0:(M-1). \tag{9}$$

The inverse transform is defined as,

$$IQDFT, f(n,m) = \frac{1}{NM} \sum_{p=0}^{N-1} W_j^{-np} \left[ \sum_{s=0}^{M-1} F_{p,s} W_k^{-ms} \right],$$
$$n = 0:(N-1), m = 0:(M-1). \tag{10}$$

## 2.4 Modified Alpha-Rooting

In the alpha-rooting method of image enhancement [4] [17] –[20] [29] –[34], for each frequency point $(p,s)$, the magnitude of the discrete quaternion transform are transformed as

$$F_{p,s} \rightarrow M[F_{p,s}]^{\alpha} \tag{11}$$

There are many different options for the choice of the M and $\alpha$ lies between $(0,1)$ [4],[17]-[20],[34]. The modified alpha rooting method is an extension of the alpha-rooting with M as a function of magnitude of the image defined as

$$M = C(p,s) = log^{\beta} \big[ |X(p,s)|^{\lambda} + 1 \big], \qquad 0 \le \beta, \qquad 0 < \lambda. \tag{12}$$

## 2.5 Color Enhancement Measure Estimation (Ceme)

The color enhancement measure estimation (CEME), is an enhancement measure [4]-[8] based on the contrast of the images. The proposed enhancement measure is related to the Weber law, which basically explains that the visual perception of the contrast is independent of luminance and low spatial frequency. To calculate the CEME value, a discrete image of size $N \times M$ is divided [21] by $k_1 k_2$ blocks of size $L_1 \times L_2$ blocks each, where $k_n = \lfloor N_n / L_n \rfloor$, for $n = 1,2$. For the RGB color model, when an image is transformed by the 2-D QDFT

$$f = (f_R, f_G, f_B) \rightarrow \hat{f} = (\hat{f}_e, \hat{f}_R, \hat{f}_G, \hat{f}_B), \tag{13}$$

where $\hat{f}_e$, referring to the scalar component of the quaternion image, which is obtained after using the transform, the CEME value is calculated by

$$E_q(\alpha) = CEME_\alpha(\hat{f}) = \frac{1}{k_1 k_2} \sum_{k=1}^{k_1} \sum_{l=1}^{k_2} 20 \log_{10} \left[ \frac{max_{k,l} \, (\hat{f}_e, \hat{f}_R, \hat{f}_G, \hat{f}_B)}{min_{k,l} \, (\hat{f}_e, \hat{f}_R, \hat{f}_G, \hat{f}_B)} \right]. \tag{14}$$

**Methodology**

The images are first enhanced by the modified alpha-rooting method. We sawin the channel-by-channel histograms of these enhanced images, that all intensity values of each of the channelwere shifted to darker regions.To increase the brightness of these enhanced images, that is, to shift the histograms back to the brighter side, the logarithmic or power-law (gamma) transformations are done. The general case of logarithmic transformations[1] is defined as

$$s = c[\log(1 + r)]^p, \tag{15}$$

where$r$ is the input image and $s$ is the resulting output image of the logarithmic transformation and $c$ is a scaling factor. The simplest case is when $p = 1$.The power law (or gamma) transformation[1] is defined as

$$s = c(r^\gamma), \tag{16}$$

Where $r$ is the input image and $s$ is the resulting output image of the gamma transformation and c is a scaling factor. When the input image itself is dark, all enhancement algorithms are performed on the negative of the image. The positive of the resultant image is taken at the final stage of image processing.

While processing with quaternion image, it is sometimes easier to see the quaternion image in the polar form. From Eq. (4), the polar form of quaternion image is $q = |q|e^{\mu_q \varphi_q}$. The spatial transformations, such as the log transformation and the gamma transformation, modify the magnitude without altering the phase of the quaternion image. If $q_{input}$ and $q_{output}$ are the quaternion image before and after applying the spatial transformation techniques, such as the log and gamma transformation techniques, then the transformed quaternion image is obtained by

$$q_{output} = |q_{output}| \frac{q_{input}}{|q_{input}|} \tag{17}$$

where$|q_{input}|$ and $|q_{output}|$ are the magnitude of the quaternion image before and after applying the transformation techniques.

## 2.6. The Enhancement Measure Estimation (EME) For Grayscale Images

The processing of the image is done on each channel separately. The 2-D DFT and the subsequently the modified alpha-rooting is done on each channel individually. The enhancement of the image by the modified alpha-rooting causes the intensity of the histogram to be shifted toward darker regions. But, further enhancement of the image by gamma transformation or logarithmic transformation results in the intensities of the histogram that are shifted towards brighter regions.

The enhancement measure estimation (EME) for grayscale images is an enhancement measure [4]-[8] based on the contrast of images. The idea behind this estimation measure is similar to that of the CEME measure for color images, but in the EME only, one color channel or grayscale image is considered. To calculate the EME value, the 2-D image of size $N \times M$ is divided[21] by $k_1 k_2$ blocks of size $L_1 \times L_2$ each.Here,$k_n = \lfloor N_n / L_n \rfloor$, for $n = 1,2$. When an image is transformed by the 2-D DFT,

$$f \rightarrow \hat{f}, \tag{18}$$

where $\hat{f}$, referring to the enhanced image, the EME value is calculated by

$$E_q(\alpha) = EME_\alpha(\hat{f}) = \frac{1}{k_1 k_2} \sum_{k=1}^{k_1} \sum_{l=1}^{k_2} 20 \, log_{10} \left[ \frac{max_{k,l}(\hat{f})}{min_{k,l}(\hat{f})} \right]. \tag{19}$$

## 3. EXPERIMENTAL RESULTS

### 3.1. Using 2-DQDFT

Figure 1is the enhanced image results of "A Streamlined Form in Lethe Vallis, Mars-esp_045833_1845[35]" was enhanced by the modified alpha rooting method. First the variation Of the CEME for $\alpha$, $\beta$, and λ values were plottedwith two parameters as variables and a fixed value for the third parameter. Figure 1a shows the surface plot of the CEME values for different values of $\alpha$, and $\beta$, for the given λ=0.58. For the image "A Streamlined Form in Lethe Vallis, Mars - esp_045833_1845", the maximum value of the CEME for the modified alpha-rooting method of enhancement is obtained at $\alpha = 0.96$, $\beta = 0.93$ for $\lambda = 0.58$.

We understood from Figure 1e that the intensity values of the histogram of the image enhanced by modified alpha-rooting method were shifted to darker regions, as compared to the histogram of the original image in Figure 1c. The CEME value obtained is higher, showing a better enhancement in the contrast. To increase the visual perception of the image, the image obtained from the modified alpha-rooting is further brightened by the logarithmic transformation. The logarithmic transformation adopted for the enhancement of the image is shown in Figure 1d, as $s = c[\log(1 + r)]^p$, with $c = 1$, and $p = 3.3$. By doing the logarithmic transformation, the histogram of the modified alpha-rooting image is shifted to brighter intensity regions.

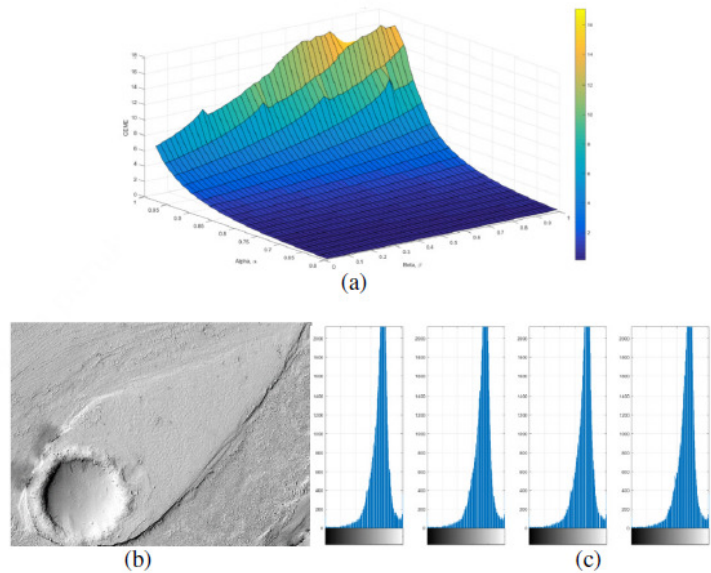

Figure1:(a) Surface plot of CEME vs $\alpha$, $\beta$ for λ=0.58 for the image "A Streamlined Form in Lethe Vallis, Mars- esp_045833_1845[35]" –¨http://www.nasa.gov/mission_pages/mars/images/index.html (b) The original image "A Streamlined Form in Lethe Vallis, Mars - esp_045833_1845¨"; (c) The histogram of the original image;

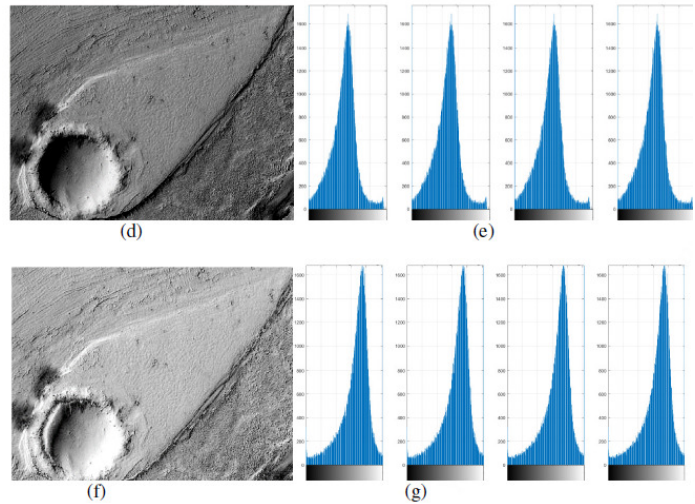

Figure1:(continued)(d) The enhanced image by the method of 2-D QDFT Modified Alpha-Rooting; (e) The histogram of the image is (d). (f) The image enhanced by the logarithmic transformation of the image in (d) and (g) the histogram of the image in (f).

The logarithmic transformation of the image enhanced by the modified alpha-rooting is also compared with the logarithmic transformation of the original image. Results of a few other algorithms were also compared side-by-side are shown in Figure 2(a) to (e).

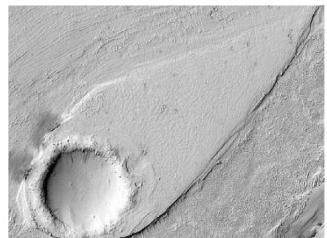

(a)

Figure 2: (a) Original Image ("A Streamlined Form in Lethe Vallis, Mars - esp_045833_1845[35]," http://www.nasa.gov/mission_pages/mars/images/index.html);

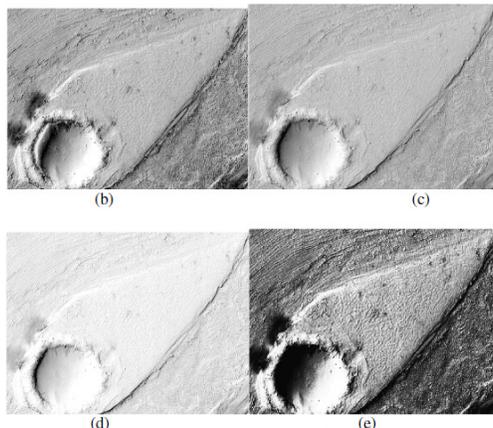

Figure 2: (continued) (b) Image enhanced by the modified alpha-rooting followed by the logarithmic transformation; (c) The color image enhancement by Retinex algorithm (McCann-99).(d) The logarithmic transformation of the original image; (e) The histogram equalization image of the original image.

It could be inferred from Figures 2a to 2d that the image enhancement by the combined effect of modified alpha rooting followed by logarithmic transformation (Figure 2b) gives a better visual perception than the original image (Figure 2a), or the enhancement by the retinex algorithm (Figure 2c) or enhancement by just the logarithmic transformation (Figure 2d). The enhanced image by modified alpha-rooting combined with logarithmic transformation, could also reveal some hidden details of the dark regions of the original image.

In the "tree" image "http://sipi.usc.edu/database/database.php?volume=misc&image=6#top", the image is first enhanced by the modified alpha-rooting and, then, the enhanced image is passed through the power law (or gamma) transformation. Enhancement results are shown in Figure 3.

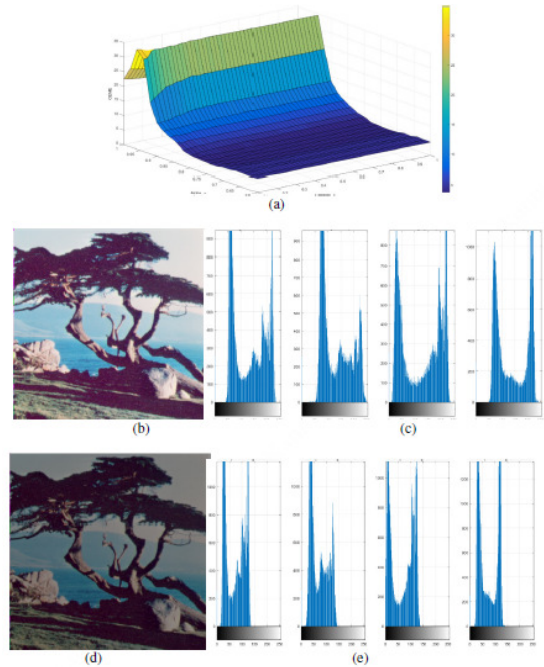

Figure 3:(a) The surface plot of the CEME vs $\alpha$ and $\lambda$ for $\beta = 0.33$ for the image "Tree[36]," - *http://sipi.usc.edu/database/database.php?volume=misc&image=6#top; (b) The original Image "tree*"; (c) the histogram of the original image; (d) the enhanced image by the method of 2-DQDFT Modified Alpha-Rooting; (e) the histogram of the image is (d);

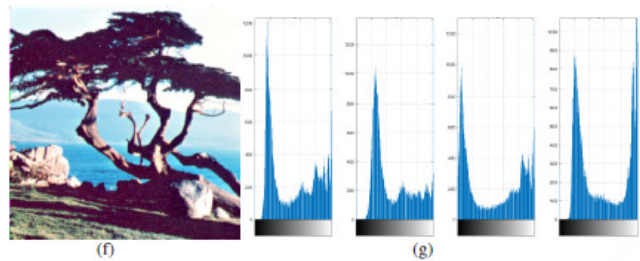

Figure3:(continued) (f) The gamma transformation of the image in (d) with gamma=1.15;(g) the histogram of the image in (f).

The "tree" image enhanced by combined enhancement method of modified alpha-rooting followed by the gamma rooting (Figure 4b) is compared with enhancement of "tree" image with the retinex algorithm (Figure 4c) and the gamma transformation method (Figure 4d).

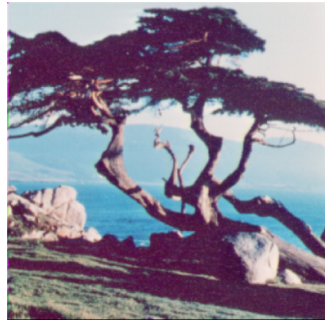

(a)

Figure 4: (a) Original image "tree[36]"-
*http://sipi.usc.edu/database/database.php?volume=misc&image=6#top ;

The comparison of images in Figure 4(a)-(e) shows that the enhancement of the image by the combined effect of modified alpha-rooting followed by the gamma rooting (Figure 4b) gave a much better enhanced image when compared to original (Figure 4a), or the enhancement by retinex method (Figure 4c), or the enhancement by the gamma transformation method alone (Figure 4d).

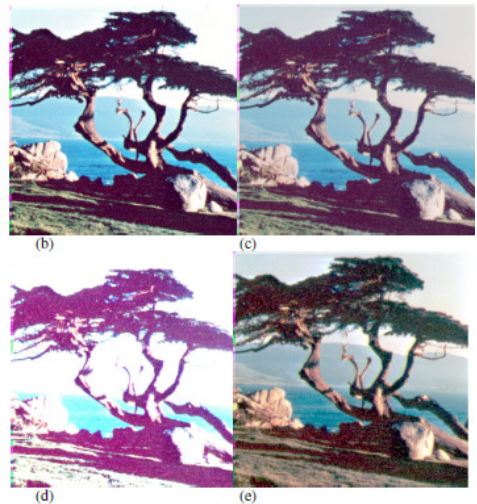

Figure 4: (continued) (b) The gamma transformation of the image enhanced by the modified alpha rooting method; (c) The color image enhancement by the retinex algorithm (McCann-99); (d) The gamma transformation alone of the original image; (e) The histogram equalized image.

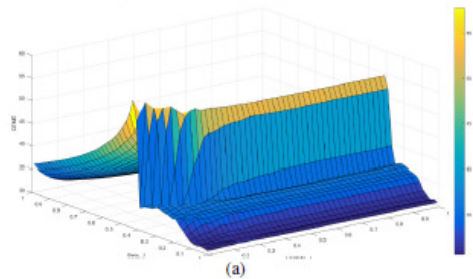

Figure5:(a) Surface plot of CEME vs $\beta$, $\lambda$ for $\alpha$=0.98 for the image "Girl in the Blue Truck – image16[37] " - *https://dragon.larc.nasa.gov/retinex/pao/news/lg-image16.jpg

In Figure 5, the original image "Girl in the Blue Truck – image16 [37]" - https://dragon.larc.nasa.gov/retinex/pao/news/lg-image16.jpg was enhanced by the modified alpha-rooting and then followed by the gamma transformation.

The value of $\beta$, and $\lambda$ are obtained from the surface plot of CEME versus $\beta$, and $\lambda$ for the given $\alpha$=0.98. The resulting enhanced images and histograms are shown in Figure 5 (b to g).

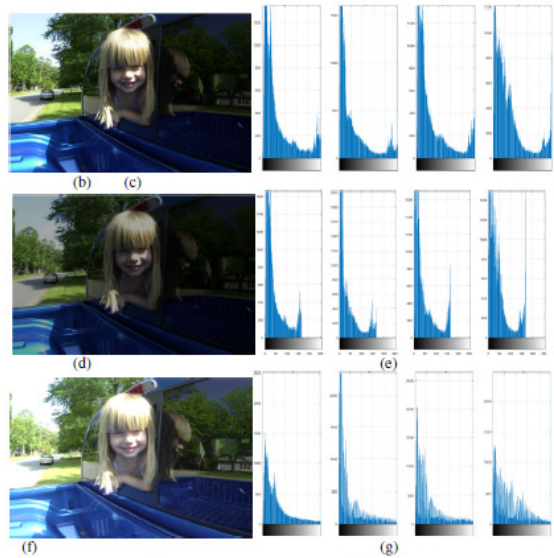

Figure 5:(continued) Surface plot of CEME vs $\beta$, $\lambda$ for $\alpha$=0.98 for the image "Girl in the Blue Truck – image16[37] " - *https://dragon.larc.nasa.gov/retinex/pao/news/lg-image16.jpg (b) The original Image "Girl in the Blue Truck – image16"; (c) The histogram of the original image; (d) The enhanced image by the method of the 2-DQDFT Modified Alpha-Rooting; (e) The histogram of the image in (d);(f) Gamma transformed image of the image in (d) for $c = 1$ and gamma=1.258; (g) histogram of the image in (f).

The enhanced image by the combined effect of the modified alpha-rooting followed by the gamma transformation can be compared with the enhanced image by the retinex algorithm and enhancement by the gamma transformation alone. The results show that the enhancement of the image "Girl in the Blue Truck – image16" by the modified alpha-rooting followed by the gamma transformation was better, than the original image, or the enhancement by the retinex algorithm, or by the gamma transformation alone.

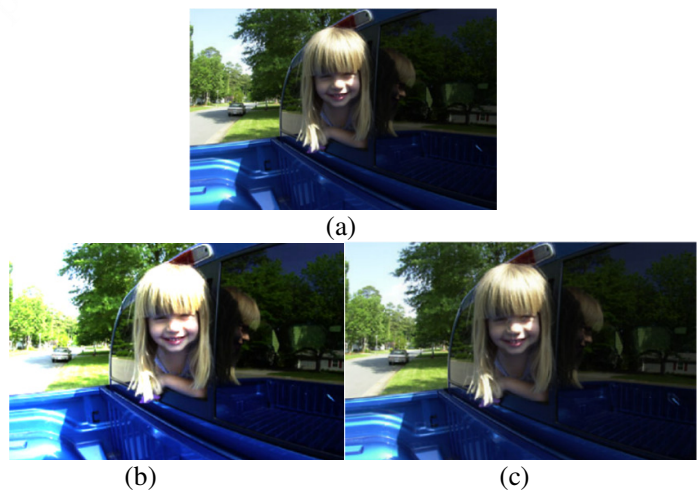

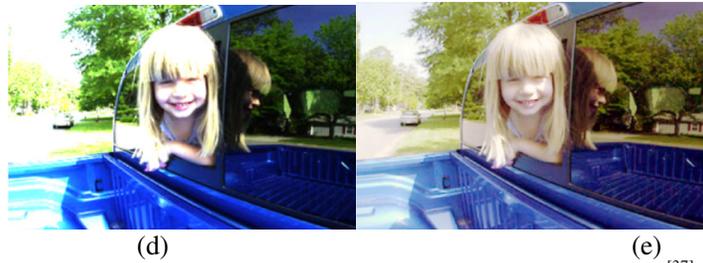

(d)                                        (e)

Figure 6: (a) The original Image "Girl in the Blue Truck – image16[37]"-
*https://dragon.larc.nasa.gov/retinex/pao/news/lg-image16.jpg ;(b) The gamma transformation of the image enhanced by the modified alpha-rooting method; (c) The color image enhancement by the retinex algorithm (McCann-99);(d) The gamma transformation alone of the original image; (e) Thehistogram equalized image of the original image.

In                    Figure7(a-e),                    "Tree"image
"http://sipi.usc.edu/database/database.php?volume=misc&image=6#top", the RGB image (Figure 7a) was first converted to the XYZ color model (Figure 7b). Then, the image enhancement techniques of the modified alpha-rooting followed by the gamma transformation were appliedon the original XYZ image (Figure 7c). The CEME measure is found to be greater, than the value obtained by the processing of the original RGB image (Figure 3f).

After processing, the image in Figure 7c was converted back to the RGB image (Figure 7d). The CEME measure of the image in Figure 7d is low as compared to the CEME measure of the image in Figure 3f, where the processing is done directly on the RGB image.

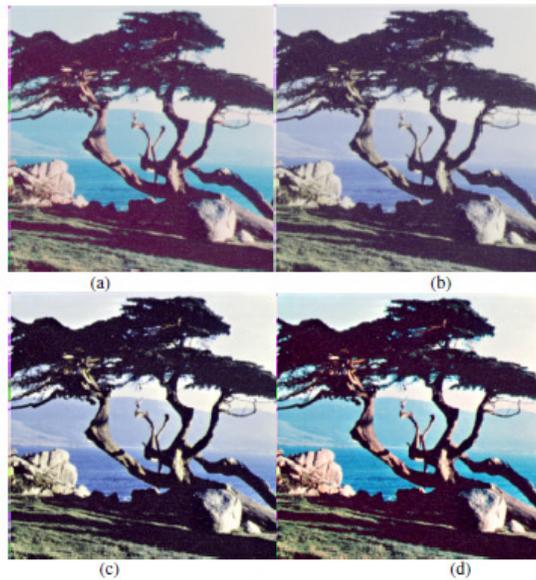

Figure 7: (a) The original RGB Image "Tree[36]"-
*http://sipi.usc.edu/database/database.php?volume=misc&image=6#top ; (b) The RGB image in (a) is converted to the XYZ color model; (c) Thegamma transformation of the image enhanced by the modified alpha rooting method done on the image in (b); (d) The image in (c) is converted back to the RGB image.

## 2.1. Using the 2-D DFT

The color image enhancement by modified alpha-rooting by the 2-D DFT of the image "Tree*"- * http://sipi.usc.edu/database/database.php?volume=misc&image=6#topwas done. The surfaces corresponding to the EME vs $\alpha$ and $\lambda$ for $\beta$ = 0.33 for all three channels were plottedseparately (Figures 8(a to c)), to find the values of $\alpha$ and $\lambda$ which give a maximum EME measure. For channel 1 (red), the maximum EME value is 35.5148 corresponding to $\alpha$=0.74 and $\lambda$=0.14. For channel 2 (green), the maximum EME value is 33.7420 corresponding to $\alpha$=0.88 and $\lambda$=0.3. For channel 3 (blue), the maximum EME value is 34.8439 corresponding to $\alpha$=0.78 and $\lambda$=0.16. The EME values of channels 1,2, and 3 of the original image equal 12.1228, 21.7077, and 12.1090, respectively.

Figure8f shows the enhanced image composed of the channels that give the maximum EME measures, when the channels were enhanced separately by the modified alpha-rooting 2-D DFT. Then, Figure8h shows the enhanced image by the combined effects of the modified alpha-rooting 2-D DFT followed by the gamma transformation with $\gamma = 3.51$.

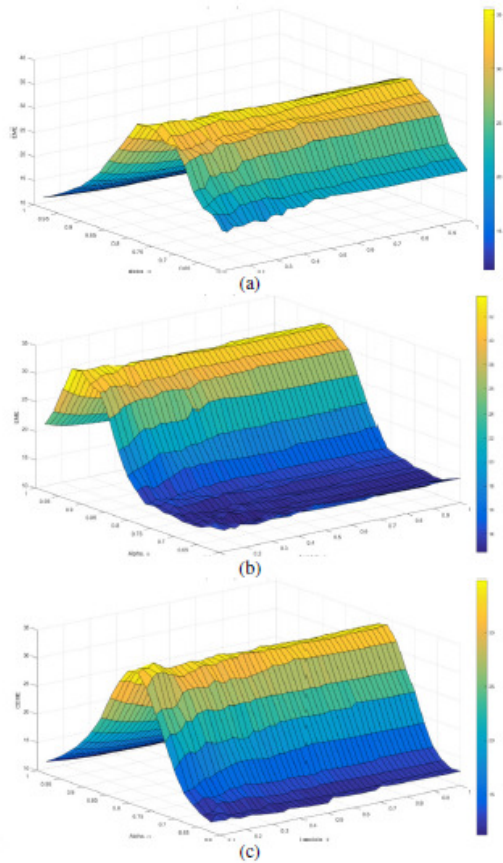

Figure 8 (a) The surface of the EME versus $\alpha$ and $\lambda$ for $\beta = 0.33$ for Channel 1 of the original image "Tree[36]"- *http://sipi.usc.edu/database/database.php?volume=misc&image=6#top ;(b) The surface of the EME versus $\alpha$ and $\lambda$ for $\beta = 0.33$ for Channel 2 for "tree" image;(c) The surface of EME vs $\alpha$ and $\lambda$ for $\beta = 0.33$ for Channel 3 for the "tree" image.;

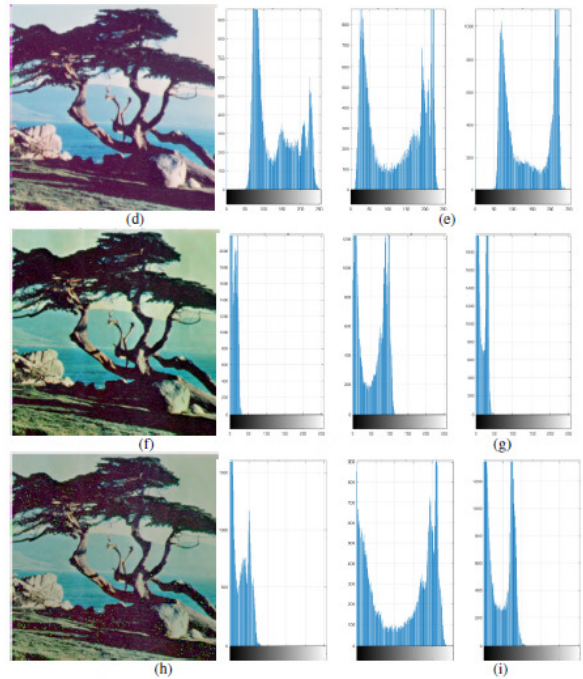

Figure 8 (continued) (c) The surface of EME vs $\alpha$ and $\lambda$ for $\beta$ = 0.33 for Channel 3 for the "tree" image.; (d) The original "Tree" image; (e) The channel by channel histogram of image in (d);(f) The channel-by-channel enhancement by the modified alpha-rooting, when the image is composed of enhanced channels corresponding to the maximum EME values; (g) The channel by channel histogram of the image in (f) before scaling;(h) The channel by channel enhancement by the modified alpha-rooting, when the image is composed of enhanced channels corresponding to the maximum EME values – followed by the gamma transformation; (i) The channel by channel histogram of image in (h) before scaling.

In Figure 9(a)-(d), the modified alpha rooting followed by the gamma transformation was accomplished (see Figure9c) in the XYZ image model (Figure9b) of the RGB image in Figure9a. After the processing the image, itwas converted back to the RGB image (in Figure9d).The $\alpha$ and $\lambda$ values for the channels X, Y, and Z corresponding to the maximums of the EME measure are respectively [0.78, 0.84, 0.78] and [0.26, 0.42, 0.16]. The EME measure of the original X, Y, and Z channels are [13.0022, 18.1088, 12.1494] and the maximums of the EME measure obtained after enhancement are [34.8012, 35.0018, 34.6950], and when the processed in the XYZ image is converted back to the RGB image, the corresponding EME measures of the channels R, G, and B are [20.8039, 36.0317, 46.7360].

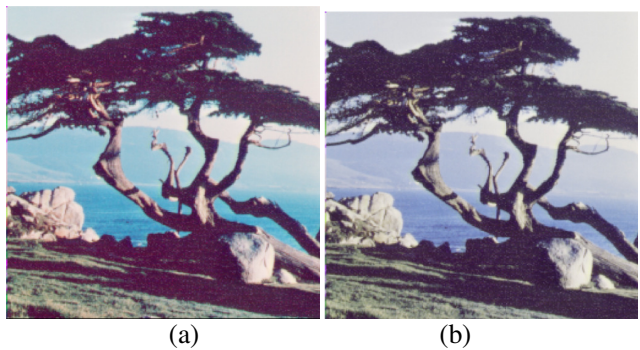

(a)                    (b)

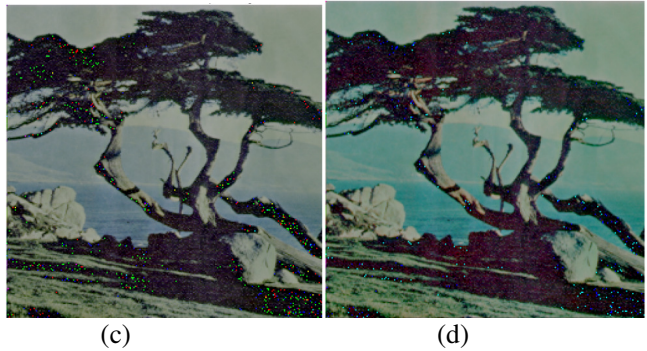

Figure 9: (a) The original RGB image "tree[62]"-
"http://sipi.usc.edu/database/database.php?volume=misc&image=6#top ; (b) The RGB image in (a) is
converted to the XYZ color model;(c) The gamma transformation of the image in (b) enhanced channel by
channel by the method of modified alpha rooting method done; (d) The image in (c) is converted back to
the RGB image.

## 4. CONCLUSIONS

The color image enhancement technique of the modified alpha-rooting by the two-side 2D quaternion discrete Fourier transform shows better enhancement techniques as compared to processing the image channel-by-channel and then composing the color image from these individual enhanced images. One of the main prominent features in channel-by-channel enhancement techniques is that the color obtained after the processing has marked difference from that of the original image and also of the color of an image when processing by the proposed quaternion approach. The integrity of the color is lost in channel-by-channel processing, while the processing in the quaternion approach preserves the uniqueness of the color. In future studies, the human sensitiveness of color perception and the psychophysics of the color will be studied better to understand the advantage of the quaternion approach of color image enhancement techniques over the channel-by-channel enhancement.

## REFERENCES

[1]  R.C. Gonzalez and R.E. Woods, Digital Image Processing, 2nd Edition, Prentice Hall, 2002.

[2]  A.M. Grigoryan, S.S. Again, A.M. Gonzales, "Fast Fourier transform-based retinex and alpha-rooting color image enhancement," [9497-29], Proc. SPIE Conf., Mobile Multimedia/Image Processing, Security, and Applications, 20 - 24 April 2015, Baltimore, Maryland United States.

[3]  A. M. Grigoryan, S.S. Agaian, "2-D Left-Side Quaternion Discrete Fourier Transform: Fast Algorithm," Proceedings, Electronic Imaging 2016, IS&T, February 14-18, San Francisco, 2016.

[4]  A.M. Grigoryan, S.S. Again, "Tensor representation of color images and fast 2-D quaternion discrete Fourier transform," [9399-16], Proceedings of SPIE vol. 9399, 2015 Electronic Imaging: Image Processing: Algorithms and Systems XIII, February 10-11, San Francisco, California, 2015.

[5]  A.M. Grigoryan and M.M. Grigoryan, Brief Notes in Advanced DSP: Fourier Analysis With MATLAB, CRC Press Taylor and Francis Group, 2009.

[6]  A.M. Grigoryan and S.S. Agaian, "Retooling of color imaging in the quaternion algebra," Applied Mathematics and Sciences: An International Journal, vol. 1, no. 3, pp. 23-39, December 2014

[7]  A.M. Grigoryan and S.S. Agaian, "Transform-based image enhancement algorithms with performance measure," Advances in Imaging and Electron Physics, Academic Press, vol. 130, pp. 165–242, May 2004.

[8]  A.M. Grigoryan and S.S. Agaian, "Optimal color image restoration: Wiener filter and quaternion Fourier transform," [9411-24], Proceedings of SPIE vol. 9411, 2015 Electronic Imaging: Mobile Devices and Multimedia: Enabling Technologies, Algorithms, and Applications 2015, February 10-11, San Francisco, California, 2015.

[9] A.M. Grigoryan, J. Jenkinson and S.S. Agaian, "Quaternion Fourier transform based alpha-rooting method for color image measurement and enhancement," SIGPRO-D-14-01083R1, Signal Processing, vol. 109, pp. 269-289, April 2015, (doi: 10.1016/j.sigpro.2014.11.019).

[10] S.S. Agaian, K. Panetta, and A.M. Grigoryan, "A new measure of image enhancement, in Proc. of the IASTED Int. Conf. Signal Processing Communication, Marbella, Spain, Sep. 1922, 2000.

[11] T.A. Ell, N. Bihan and S.J. Sangwine, "Quaternion Fourier transforms for signal and image processing," Wiley NJ and ISTE UK, 2014

[12] W.R. Hamilton, Elements of Quaternions, Logmans, Green and Co., London, 1866.

[13] T.A. Ell, "Quaternion-Fourier transforms for analysis of 2-dimensional linear time-invariant partial differential systems," In Proceedings of the 3nd IEEE Conference on Decision and Control, vol. 1-4, pp. 1830–1841, San Antonio, Texas, USA, December 1993.

[14] S.J. Sangwine, "Fourier transforms of colour images using quaternion, or hypercomplex, numbers," Electronics Letters, vol. 32, no. 21, pp. 1979–1980, October 1996.

[15] S.J. Sangwine, "The discrete quaternion-Fourier transform," IPA97, Conference Publication no. 443, pp. 790–793, July 1997.

[16] S.J. Sangwine, T.A. Ell, J.M. Blackledge, and M.J. Turner, "The discrete Fourier transform of a color image," in Proc. Image Processing II Mathematical Methods, Algorithms and Applications, pp. 430–441, 2000.

[17] S.J. Sangwine and T.A. Ell, "Hypercomplex Fourier transforms of color images," in Proc. IEEE Intl. Conf. Image Process., vol. 1, pp. 137-140, 2001

[18] S.S. Agaian, K. Panetta, and A.M. Grigoryan, (2001) "Transform-based image enhancement algorithms," IEEE Trans. on Image Processing, vol. 10, pp. 367-382.

[19] A.M. Grigoryan and S.S. Agaian, (2014) "Alpha-rooting method of color image enhancement by discrete quaternion Fourier transform," [9019-3], in Proc. SPIE 9019, Image Processing: Algorithms and Systems XII, 901904; 12 p., doi: 10.1117/12.2040596

[20] J.H. McClellan, (1980) "Artifacts in alpha-rooting of images," Proc. IEEE Int. Conf. Acoustics, Speech, and Signal Processing, pp. 449-452.

[21] F.T. Arslan and A.M. Grigoryan, (2006) "Fast splitting alpha-rooting method of image enhancement: Tensor representation," IEEE Trans. on Image Processing, vol. 15, no. 11, pp. 3375-3384.

[22] A.M. Grigoryan and F.T. Arslan, "Image enhancement by the tensor transform," Proceedings, IEEE International Symposium on Biomedical Imaging: from Macro to Nano, ISBI 2004, pp. 816-819, April 2004, Arlington, VA.

[23] A.M. Grigoryan and S.S. Agaian, "Cell Phone Cameral Color Medical Imaging via Fast Fourier Transform," in Electronic Imaging Applications in Mobile Healthcare, J. Tang, S.S. Agaian, and J. Tan, Eds., SPIE Press, Bellingham, Washington, pp. 33-76 (2016).

[24] A.M. Grigoryan and S. Dursun, "Multiresolution of the Fourier transform," in the Proceedings of the IEEE International Conference on Acoustics, Speech, and Signal Processing, 2005 (ICASSP '05), vol. 4, pp. 577–580, March 18-23, 2005, Philadelphia, PA

[25] A.M. Grigoryan and S.S. Agaian, "Tensor form of image representation: Enhancement by image-signals," [5014-26] (Image Processing: Algorithms and Systems II, Electronic Imaging, Science and Technology 2003, IS&T/SPIE 15 Annual Symposium, Santa Clara, CA, January 20-24, 2003)

[26] A.M. Grigoryan, "2-D and 1-D multi-paired transforms: Frequency-time type wavelets," IEEE Trans. on Signal Processing, vol. 49, no. 2, 344–353 (2001)

[27] S.S. Agaian, H.G. Sarukhanyan, K.O. Egiazarian, J. Astola, Hadamard transforms, SPIE Press, 2011.

[28] A.M. Grigoryan, K. Naghdali, "On a method of paired representation: Enhancement and decomposition by series direction images," Journal of Mathematical Imaging and Vision, vol. 34, no. 2, 185–199 (2009)

[29] F.T. Arslan and A.M. Grigoryan, "Enhancement of medical images by the paired transform," in Proceedings of the 14th IEEE International Conference on Image Processing, vol. 1, pp. 537-540, San Antonio, Texas, September 16-19, 2007.

[30] F.T. Arslan, J.M. Moreno, and A.M. Grigoryan, "Paired directional transform-based methods of image enhancement," in Proceedings of SPIE – vol. 6246, Visual Information Processing XV, April 18-19, 2006.

[31] A.M. Grigoryan and S.S. Agaian, "Color Enhancement and Correction for Camera Cell Phone Medical Images Using Quaternion Tools," in Electronic Imaging Applications in Mobile Healthcare, J. Tang, S.S. Agaian, and J. Tan, Eds., SPIE Press, Bellingham, Washington, pp. 77-117 (2016).

[32] A.M. Grigoryan, S.S. Agaian, "Alpha-Rooting Method of Gray-scale Image Enhancement in The Quaternion Frequency Domain," Proceedings of IS&T International Symposium, Electronic Imaging: Algorithms and Systems XV, 29 Jan.-2 Feb., Burlingame, CA, 2017.

[33] A.M. Grigoryan, A. John, S.S. Agaian, "Color image enhancement of medical images using alpha-rooting and zonal alpha-rooting methods on 2-D QDFT," Proceedings of SPIE Medical Imaging Symposium, Image Perception, Observer Performance, and Technology Assessment, 12 - 13 Feb., Orlando, FL, 2017.

[34] http://www.nasa.gov/mission_pages/mars/images/index.html

[35] http://sipi.usc.edu/database/database.php?volume=misc&image=6#top

[36] https://dragon.larc.nasa.gov/retinex/pao/news/lg-image16.jpg

## AUTHORS

**Artyom M. Grigoryan** received the PhD degree in Mathematics and Physics from Yerevan State University (1990). He is an associate Professor of the Department of Electrical Engineering in the College of Engineering, University of Texas at San Antonio. The author of four books, 9 book-chapters, 3 patents and more than 120 papers and specializing in the design of robust filters, fast transforms, tensor and paired transforms, discrete tomography, quaternion imaging, image encryption, processing biomedical images. 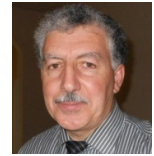

**Aparna John** received her B. Tech. degree in Applied Electronics and Instrumentation from University of Calicut, India and M. Tech. degree in Electronics and Communication with specialization in Optoelectronics and Optical Communication from University of Kerala, India. Now, she is a doctoral student in Electrical Engineering at University of Texas at San Antonio. She is pursuing her research under the supervision of Dr. Artyom M. Grigoryan. Her research interests include image processing, color image enhancements, fast algorithms, quaternion algebra and quaternion transforms including quaternion Fourier transforms. She is a student member of IEEE and also a member of Eta Kappa Nu Honor Society, University of Texas San Antonio 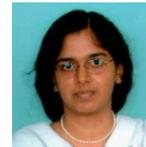

**Sos S. Agaian** is the Distinguished Professor at the City University of New York/CSI.  Dr. Agaian received the Ph.D. degree in math and physics from the Steklov Institute of Mathematics, Russian Academy of Sciences, and the Doctor of Engineering Sciences degree from the Institute of the Control System, Russian Academy of Sciences. He has authored more than 500 scientific papers, 7 books, and holds 14 patents.  His research interests are Multimedia Processing, Imaging Systems, Information Security, Artificial Intelligent, Computer Vision, 3D Imaging Sensors, Fusion, Biomedical and Health Informatics. 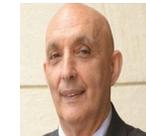